\documentclass[sigconf]{acmart}

\usepackage{amsmath}
\usepackage{booktabs}
\usepackage{graphicx}
\usepackage{color}
\usepackage{xcolor}
\usepackage{multirow}

\usepackage[all]{foreign}
\usepackage{subfig}
\usepackage{algorithm}
\usepackage{algorithmic}
\usepackage{hhline}
\usepackage{soul}
\sethlcolor{white}
\usepackage{wrapfig}
\usepackage{multirow}
\usepackage[inline]{enumitem}
\AtBeginDocument{%
  }

\setcopyright{acmcopyright}
\copyrightyear{2025}
\acmYear{2025}
\acmDOI{XXXXXXX.XXXXXXX}

\acmJournal{JACM}
\acmVolume{37}
\acmNumber{4}
\acmArticle{111}
\acmMonth{8}

\newcommand{\ebsc}{\textit{ElectionBot-SC} }
\newcommand{\sca} {\textit{SafeChat} }

\newcommand{\nobulletline}[1]{\noindent {\bf #1}}

\begin{document}

\title{SafeChat: A Framework for Building Trustworthy Collaborative Assistants and a Case Study of its Usefulness}


\author{Biplav Srivastava}
\orcid{0000-0002-7292-3838}
\affiliation{%
  \institution{University of South Carolina}
  \streetaddress{1112 Greene St}
  \city{Columbia}
  \state{South Carolina}
  \country{USA}}
\email{biplav.s@sc.edu}

\author{Kausik Lakkaraju}
\orcid{0000-0002-7292-3838}
\affiliation{%
  \institution{University of South Carolina}
  \streetaddress{1112 Greene St}
  \city{Columbia}
  \state{South Carolina}
  \country{USA}}
\email{kausik@email.sc.edu}

\author{Nitin Gupta}
\orcid{0009-0001-7518-3994}
\affiliation{%
  \institution{University of South Carolina}
  \streetaddress{1112 Greene St}
  \city{Columbia}
  \state{South Carolina}
  \country{USA}}
\email{niting@email.sc.edu}

\author{Vansh Nagpal}
\orcid{0000-0003-0605-9057}
\affiliation{%
\institution{University of South Carolina}
  \streetaddress{1112 Greene St}
  \city{Columbia}
  \state{South Carolina}
  \country{USA}}
\email{vnagpal@email.sc.edu}

\author{Bharath C. Muppasani}
\orcid{0000-0003-3782-8672}
\affiliation{%
  \institution{University of South Carolina}
  \streetaddress{1112 Greene St}
  \city{Columbia}
  \state{South Carolina}
  \country{USA}}
\email{bharath@email.sc.edu}

\author{Sara E. Jones}
\orcid{0009-0008-9208-9933}
\affiliation{%
  \institution{University of South Carolina}
  \streetaddress{1112 Greene St}
  \city{Columbia}
  \state{South Carolina}
  \country{USA}}
\email{sej15@email.sc.edu}







\renewcommand{\shortauthors}{Srivastava et al.}

\begin{abstract}


Collaborative assistants, commonly known as assistants or chatbots, are data-driven decision support systems that enable natural user interaction for task completion. While they have the potential to meet critical informational needs in modern society, concerns persist about their reliability and trustworthiness. In particular, Large Language Model (LLM) based chatbots such as ChatGPT, Gemini, and DeepSeek, are becoming very accessible. However, there are many issues with such chatbots
including their inability to explain how they  generated their responses, risk of exhibiting 
problematic content like abusive language,  absence of   standardized testing procedures to assess for acceptable risk and ensure their reliability, and the need for  deep Artificial Intelligence (AI) expertise and extended development times for building them. As a result, chatbots have been currently considered inappropriate for trust-sensitive applications like elections or health. To overcome these
issues, we introduce {\em SafeChat}, a general architecture  designed to build chatbots that are safe and trustworthy, {\em expeditiously}, with a particular focus on information retrieval usecases. Its key capabilities include: (a) for {\bf safety}, a domain-agnostic, safe-design where only responses that are grounded and traceable to an allowed source will be answered ({\em provenance}) and provision for {\em do-not-respond} strategies that can deflect certain user questions which may be harmful if answered; (b) for {\bf usability}, automatic, extractive summarization of long answers that can be traced back to a source and automated trust assessment  to communicate the chatbot’s expected behavior on dimensions like sentiment; and (c)  for {\bf fast, scalable, development}, a CSV-driven workflow,  provision for automated testing, and integration with a range of devices. We have implemented \sca into an executable framework using an open-source chatbot building platform, Rasa, and discuss a case study of using it to build an effective chatbot, \ebsc,  for safely disseminating official information for elections. \sca is being used in many domains by us and others validating its general potential, and is  available at: \url{https://github.com/ai4society/trustworthy-chatbot}.

\end{abstract}





\received{20 February 2007}
\received[revised]{12 March 2009}
\received[accepted]{5 June 2009}

\maketitle

\section{Introduction}


All organizations are holders of  authentic information - whether about themselves, like businesses about their products and services, or about others, like public agencies mandated to record information about individuals, organizations and civic services.
There are many use cases where organizations need to disseminate these {\em official} content  to stakeholders - customers, residents, suppliers, and employees - in easy to understand forms. 
Ensuring that official information is accurately conveyed is crucial for informed decision-making at both individual and community levels.
With the growing availability of online Artificial Intelligence (AI) services, such as search engines and now, collaborative assistants -  chatbots or assistants for short, particularly those powered by Large Language Models (LLMs), users have more options for accessing information than ever before.

However, these  advanced AI services are neither trustworthy nor easy to build. 
Despite their convenience, real-world testing in critical scenarios, such as the 2024 elections \cite{openai-chatgpt-elections,srivastava2025vision}, finance \cite{finance-llm-icaif} or public health crises like COVID-19  \cite{apollo-chatbots}, have shown that chatbots often fail to meet the necessary standards for reliability and effectiveness - including hallucinations \cite{rawte-etal-2023-hallucination-emnlp}.
Studies show that these assistants can have error rates exceeding 50\%, with many mistakes going unnoticed by users, leading to frustration and diminished trust \cite{maharaj-etal-2024-chatbot-evaluation-adobe}.
The issues can be summarized as their inability to explain how they  generated their responses, risk of exhibiting 
problematic responses like abusive language or harm,  absence of   standardized testing procedures to assess for acceptable risk and ensure their reliability\cite{srivastava2023evaluatingchatbotspromoteusers}, and the need for  deep Artificial Intelligence (AI) expertise and extended development times for building them.
The current state-of-affairs is ironical considering
that chatbots have been studied as decision support tools since the early days of AI and they have gained unprecedented public and commercial interest with the introduction of user-friendly, general-purpose LLM-based chatbots like ChatGPT, Gemini, and DeepSeek. 
Thus, there is a growing need for  improvement in chatbot development, performance and reliability. If these issues remain unaddressed, the AI community risks missing a crucial opportunity to meet the informational needs of businesses and society effectively.

In response, we introduce {\em SafeChat}, a general architecture designed to build chatbots that are safe and trustworthy quickly with a particular focus on information retrieval. Its key capabilities include: (a) for {\bf safety}, a domain-agnostic, safe-design where only responses that are grounded and traceable to an allowed source will be answered and provision for {\em do-not-respond} strategies that can deflect certain user questions which may be harmful if answered; (b)
for {\bf usability}, automatic, extractive summarization of long answers that can be traced back to a source, and automated trust assessment  to communicate the chatbot’s expected behavior on dimensions like sentiment; and (c)  for {\bf fast, scalable, development}, a CSV-driven workflow, and provision for automated testing and integration with a range of devices. We implement \sca using a leading open-source, rule-based chatbot building platform, Rasa\cite{rasa}, and discuss a case study of using the tool to build an effective chatbot for safely disseminating official information for elections.
\sca itself has been used in many domains by us and others validating its general potential. In the presented context, \sca is also well suited for official sources to 
{\bf spread reliable information in a user-personalized and consumable content format in terms of
language, structure and delivery}. The code is available publicly\cite{safechat-arch-github}.

In the remainder of the paper, 
we will first present the collaborative information retrieval problem tackled and then our  \sca framework. Next, we discuss the use case 
of disseminating official election information,  how \sca was used to build \ebsc for South Carolina, and its evaluation during the 2024 elections. 
We conclude the paper with a discussion of how \sca is handling risks as per an emerging AI risk guideline, NIST Risk Management Framework \cite{nist-rmf}, how it is being used more widely, and some future research directions.



\section{Context and  the (Collaborative IR) Problem Addressed}

We consider a common scenario of collaboration between a user and a system where the former seeks  information on a subject and the latter provides it. Although this can be  served with an information retrieval  (IR) system when the user {\bf knows the informational need} (e.g., query) well, as analyzed in \cite{conversation-ir-framework}, a longer human-system conversations   are more apt when their roles are to  (a) elicit actual user needs by helping them formulate queries clearly, and (b) helping users when there need is satisfied  with a set of results that interact to produce a single item response, like a travel itinerary involving a flight, hotel stay and a car rental. The authors identify five patterns employing these roles - \textbf{(1)} the chatbot system helps the user reveal their needs, \textbf{(2)} the system itself reveals its capabilities, \textbf{(3)} either the system or the user can initiate the conversation, \textbf{(4)} the system has memory and recalls what user may need, and \textbf{(5)} the system can reason about the utility of sets
of complementary items, to create a response.

Motivated by the IR conversation scenario, we  now describe our problem with its inputs and outputs. Let $T$ be the set of all text strings and $U \subseteq T$ be the set of all user input utterances to the system. Additionally, we introduce the dataset $D$, comprising \textit{domain-specific} frequently asked questions (FAQs) ($D^{{S_j}}_{FAQ}$), \textit{domain-independent} FAQs list ($D^{I}_{FAQ}$), 
and questions to be avoided ($D^A$). Thus, $D = D^{{S_j}}_{FAQ} \cup D^{I}_{FAQ} \cup D^A$. 

Here, $ D^{{S_j}}_{FAQ}=\{\{q_1,a_1\}, \{q_2,a_2\}, ..., \{q_m,a_m\}\}$, where each ($q_i, a_i$) is a pair of question and its corresponding answer  within a specific domain ($S_j$) such as elections, finance, or public health. Similarly, $D^{I}_{FAQ} = \{\{q_1,a_1\}, \{q_2,a_2\}, ..., \{q_n,a_n\}\}$ comprises generic questions and answers, such as initiating or concluding dialogues (greetings, chitchat, etc.). We also incorporate a set of queries  requiring deflection  based on {\em do-not-answer} (DNA) strategies denoted as $D^A$. The DNA response strategy is a  {\em a set} of responses with associated probabilities, i.e.,  $D^A = \{(q_k, A_k)\}$ where  $A_k = \{(a_k, p_k) \}$. Thus, each $ a_k$ is an answer deflecting the question like a null response (e.g. \textit{I am unable to answer that question}) with probability $p_{\bot}$, a humor response (e.g. a winking emoji) with probability $p_h$, or an alternative question (e.g. \textit{Did you mean to ask ...?}) with probability $p_{a_i}$. The probabilities $p_{\bot}, p_h, p_{a_i}$, .. can be configured at the system-level or dynamically adjusted, and must sum to 1. This approach facilitates user query deflection by providing a direct means to get an answer or redirecting to another relevant question for a suitable response. 
Furthermore, each answer, $a_i$, can be a static answer (string) or a result from {\em a known function} (e.g., REST - Representational State Transfer - API) ) call, $\omega_i({q_i})$,  to a get the latest value of a dynamic phenomenon (e.g., temperature). 


The output of the chatbot is a function $f: U \rightarrow R$, where for any user input utterance $u_i \in U$, $f(u_i) = r_i \in R$, where $R = \{a_i| (q_i, a_i) \vee \omega_i({q_i})  \in D\}$ and $\delta(u_j, q_i) \leq \epsilon$. 
Here, $\delta$ is a distance function and $\epsilon$ is an acceptance threshold.

Our formulation allows the chatbot designer to be in control of the answers, support dynamic answers via REST APIs, reuse common responses for domain independent utterances, and when deciding not to answer, maintain curiosity by choosing answers stochastically. The user may pose any utterance and the system can learn to improve its detection of user intent (query) from data.


\section{Towards a Solution and Implementation}

\begin{figure*}
	\includegraphics[width=0.7\textwidth]{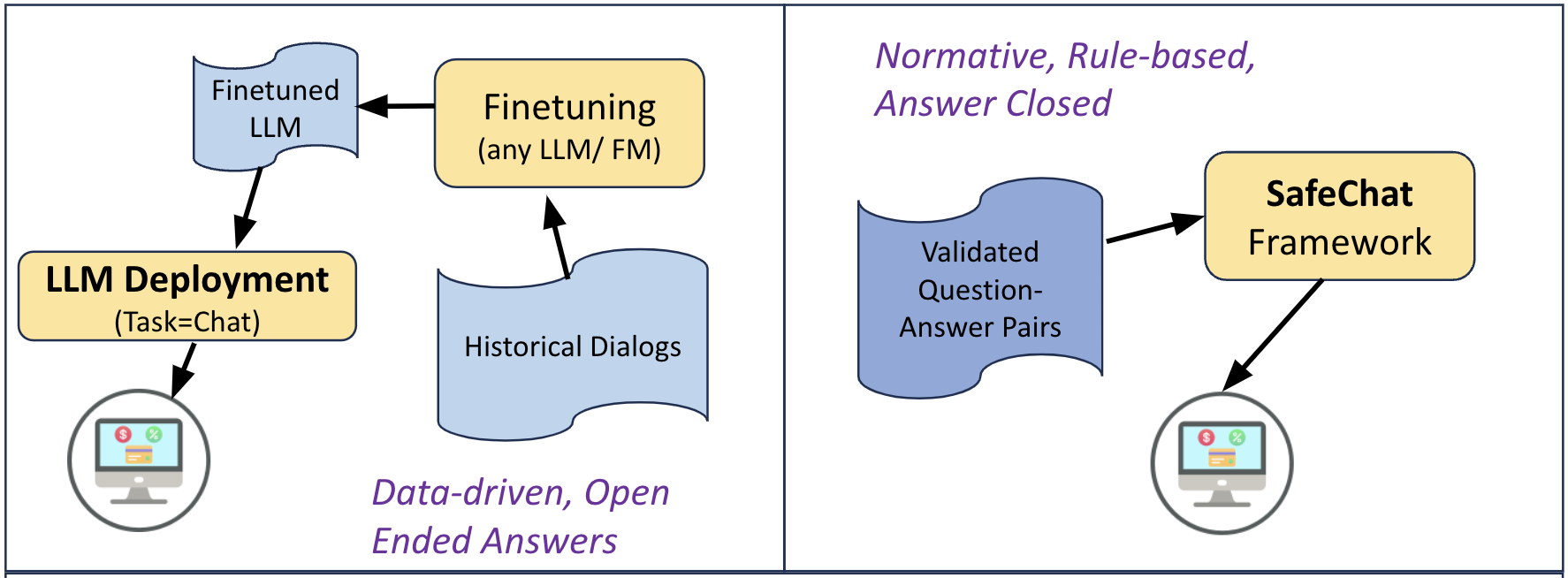}
	\caption{Comparing chatbot approaches: (left) a data-driven, open-ended LLM approach, where an LLM is finetuned on historical dialogs for chat-based tasks; and (right) a normative, rule-based SafeChat framework built on validated question-answer pairs for controlled, closed-ended responses.}
	\label{fig:chatbotapproaches}
\end{figure*}

We now see how a LLM-based chatbot may operate on the problem and then introduce the \sca approach with features to address our problem. As illustrated in Figure~\ref{fig:chatbotapproaches} (left), an LLM is trained on a general corpus and in order to specially excel in a domain, must be finetuned on historical dialogs from it. Then, it can be deployed for the {\em chat} task for generating any response to user's utterance.
In contrast, \sca focuses on validated data from the very start Figure~\ref{fig:chatbotapproaches} (right) and with the goals of promoting safety, usability, and rapid development, builds  capabilities that can be used along with  any rule-based platform to create chatbots.

\begin{figure*}
	\includegraphics[width=0.8\textwidth]{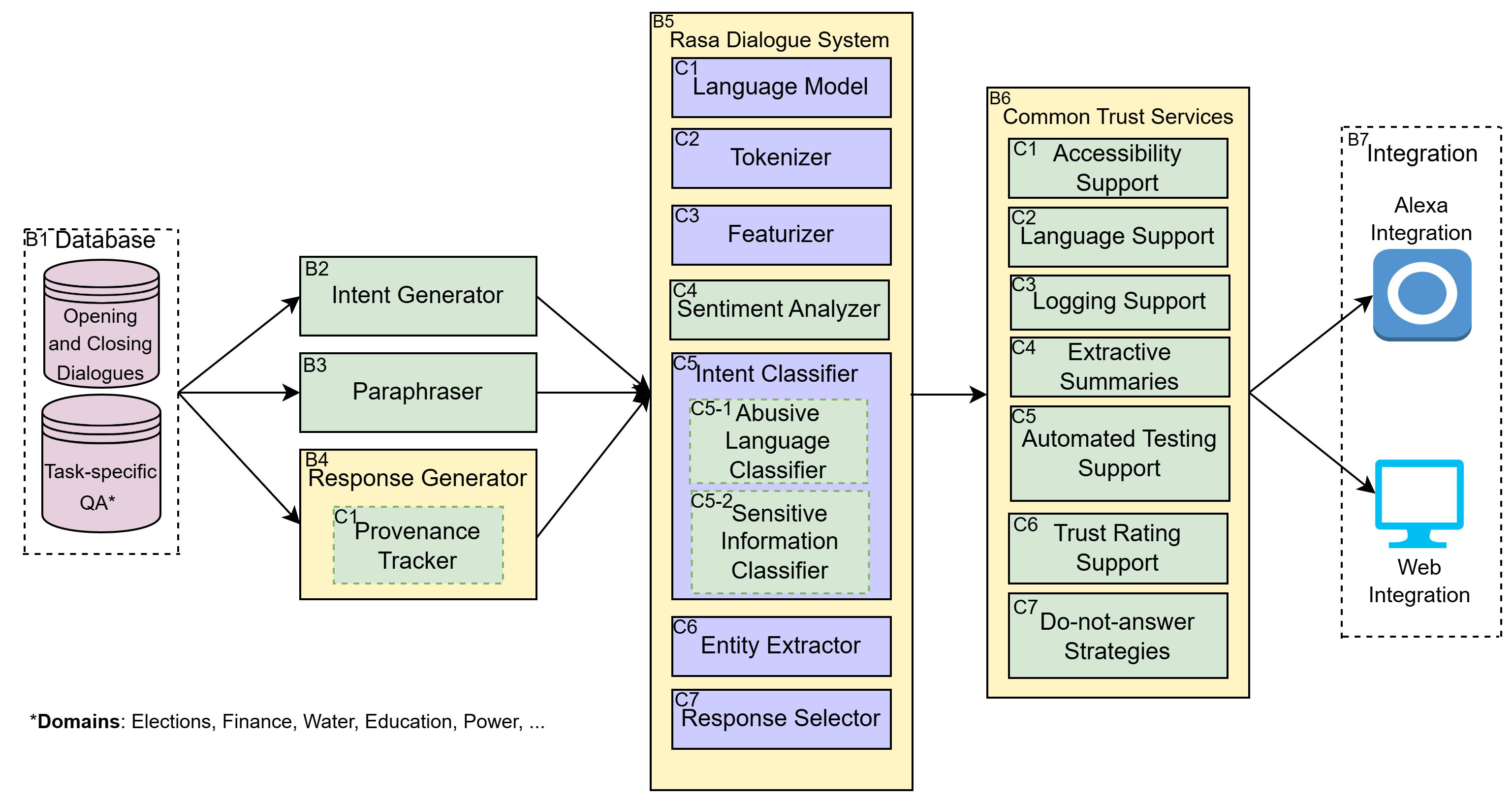}
	\caption{The \sca System Architecture. Novel components are B4-C1 (provenance tracker), B5-C4 (sentiment analyzer), B5-C5 (Intent Classifier), and all sub-components of B6 (i.e., B6-C1 to B6-C7). Other components are leveraged from RASA.}
	\label{fig:sys-arch}
\end{figure*}

We have chosen  the RASA chatbot framework \cite{rasa} to implement the \sca architecture and explain it first before going into the latters components. 
RASA incorporates a Natural Language Understanding (NLU) pipeline designed to interpret and respond to human input effectively. The system can integrate language models such as the Spacy model for utilizing pre-trained word vectors. The tokenizer converts sentences into tokens, while the featurizer generates vector representations of both the user’s message and the response. The intent classifier determines the underlying intent of the user’s input, while the entity extractor identifies any relevant entities embedded within the query. Finally, the response selector, based on  the identified intent and extracted entities, chooses the most appropriate response according to pre-specified rules.
\sca is shown in Figure ~\ref{fig:sys-arch} and we describe its components  next.

\nobulletline{Database (B1)}: 
The database is the source from which we obtain the training data to train the chatbot. We assume that the user has chosen the data source(s) that are reliable and trustworthy which in-turn will be used for training the intent detectors that recognize user utterances. Task-specific QA refers to the data source pertaining to the chosen domain (e.g., elections, health). The opening and closing dialogues are reusable, such as greetings and farewells.

\nobulletline{Intent Generator (B2)}: 
Intent Generation is crucial in building an efficient and functional chatbot that will determine its success in fulfilling the user's needs. Intent Generator helps in tagging existing questions to an intent, which can later be utilized to map any new incoming user utterance to an available intent to provide desired answers. Given the domain-specific data, $D^{{S_j}}_{FAQ}$, we generate unique intents for each question automatically by creating n-grams (i.e., $n$ word sequences), starting with the most specific n-gram (longest combination of words; n=5 by default) that best represents the intent removing stopwords and adjusting for query length. 
We also resolve duplicate intents by modifying them to ensure each query has a unique label.


\nobulletline{Paraphraser (B3)}: In order to train a chatbot, it is crucial to present it with similar questions that match an answer, rather than having only a single question associated with an answer. However, in any FAQ data, we only obtain a single question tagged with an answer. Thus, in order to augment the training data to build an efficient chatbot, we use a pre-trained T5-based paraphrasing model to generate multiple paraphrases for each input question by tokenizing the question and then decoding the generated tokens into paraphrases.  

\nobulletline{Response Generator (B4)}: Responses are messages that the chatbot sends to the user. A response is usually text, but can also include multi-modal content like images and audio. \sca reuses the default response generation module available in the RASA Dialogue System. This module helps provide various ways to respond to a user's utterance using Custom Actions, Response Variations, Conditional Response Variations, Images, and Rich Response Buttons among many others.
    \textbf{Provenance Tracker (B4-C1)}: Provenance tracker is responsible for linking the output generated by the chatbot to its original source which is provided in the input CSV file as a column. The source can be a static website from which the response is extracted or a REST endpoint.

\nobulletline{Dialog System (B5)}: We extend the dialog system's (i.e., RASA's) handling of user utterance with specific new capabilities.
\begin{itemize}
    \item \textbf{Sentiment Analyzer (B5-C4)}: It estimates the sentiment value of the user message. It uses the Textblob sentiment analyzer to estimate the semantic orientation (positive, negative, or neutral) and also estimates the magnitude of the sentiment.
    \item \textbf{Abusive Language Classifier (B5-C5)}: It checks the user messages for any abusive language.
    \item \textbf{Sensitive Information Classifier (B5-C6)}: It checks if one user is asking for any sensitive/personal information about other users.
\end{itemize}

\nobulletline{Common Services (B6)}: The common services are optional and the \sca instance has the flexibility of choosing the services they need. 
\begin{itemize}
\item \textbf{Accessibility Support (B6-C1)}: This includes font settings and Text-to-Speech capability. 
\item \textbf{Language Support (B6-C2)}: The users will be able to converse with the chatbot in a language that is comfortable. 
\item \textbf{Logging Support (B6-C3)}: The conversations can also be logged for storage and retrieval. This can help the developers improve the chatbot conversation by reviewing the stored conversations. 

\item \textbf{Extractive Summaries (B6-C4)}: Some answers can be long in the official FAQs. So we let the user decide if they want an {\em extractive summary} of the original answer, which returns text that is only from the validated answers. 

\item \textbf{Automated Testing Support (B6-C5)}: Provides support for testing. Currently, we support  recording of feedback from user (for surveys), but in future, we plan to support randomized control trials (RCT) by enabling creation of chatbot instances  for control and treatment groups, and analysis of results. 

\item \textbf{Trust Rating Support (B6-C6)}: This can be used assess and communicate the chatbot’s expected behavior on dimensions like sentiment (supported), abusive language and bias \cite{kiritchenko2021confronting,srivastava2020personalized}. 
\item \textbf{Do-not-answer (DNA) Strategies (B6-C7)}: This is meant for handling cases that the chatbot designer considers sensitive or harmful, and wants to avoid. The chatbot would recognize the intent  and {\em purposefully} not answer. e.g., hypothetical questions not traceable to source data - ‘Whom should I vote for?’. DNA corresponds to $D^A$.

\end{itemize}
\nobulletline{System Integration}: Integrating a chatbot with user-friendly hardware increases their accessibility for users.
We rely on  RASA's integration features for the same. In particular, for Alexa devices, we have created a {\em skill} using the Alexa developer console (beta phase). 

The \sca design has been informed by our experience in building chatbots in education \cite{allure} and  home automation \cite{nl2sql}). The presented framework has been used for  new domains like election (case study presented), finance,  medical student training), and library services demonstrating the generalizability of our framework. 



\section{Case Study: Chatbot for Disseminating Official Elections Information}

To improve usability of official election information in South Carolina (SC), a US state, for potential voters, we developed an instance of \sca called \ebsc.
We now describe its details and its effectiveness in the 2024 elections.


\subsection{Dataset}
To effectively train \sca on a domain, one needs to compile a dataset of question-answer pairs that cover the domain's topics. 
For SC, 
we compiled FAQs from the website of SC election commission  \cite{sc-faqs} as the primary source, and since its size was small, complemented it with data  from a non-profit, the League of Women Voters \cite{vote411}, as the secondary source. 
Our dataset evolved from an initial version in October $2022$ with $30$ QA pairs covering $10$ topics to an enhanced version in September $2024$ featuring $23$ QA pairs across $8$ topics. We retained $7$ QA pairs from the older version that were missing, but relevant, in the update. Additionally, we incorporated $11$ QA pairs from Vote411 to broaden coverage. The final dataset comprises $41$ QA pairs covering $9$ topics, enhancing official state information with reputable non-governmental sources. Table \ref{tab:data_statistics} presents key statistics of the FAQ dataset, including QA pair counts, average question and answer lengths, and topic coverage across sources.

\begin{table}[!b]
    \centering
    \footnotesize
    \begin{tabular}{l|c|c|c|c}
    \toprule
    \textbf{Metric} & \multicolumn{2}{c|}{\textbf{South Carolina}} & \textbf{Vote411} & \textbf{Used} \\
    \midrule
    \# QA pairs & 30 & 23 & 11 & 41 \\
    Avg. question length & 10.9 & 7.6 & 14.5 & 11.9 \\
    Avg. answer length & 80.9 & 51.3 & 80.9 & 70.6 \\
    \# Topics & 10 & 8 & 11 & 9 \\
    Last updated & Oct 2022 & Sep 2024 & Sep 2024 & Sep 2024 \\
    \bottomrule
    \end{tabular}%
    \caption{Statistics about FAQ datasets. Question and answer lengths are measured in words.}
    \label{tab:data_statistics}
\end{table}

\subsection{User Interface and Interactivity}

\begin{figure*}
	\includegraphics[width=0.8\textwidth]{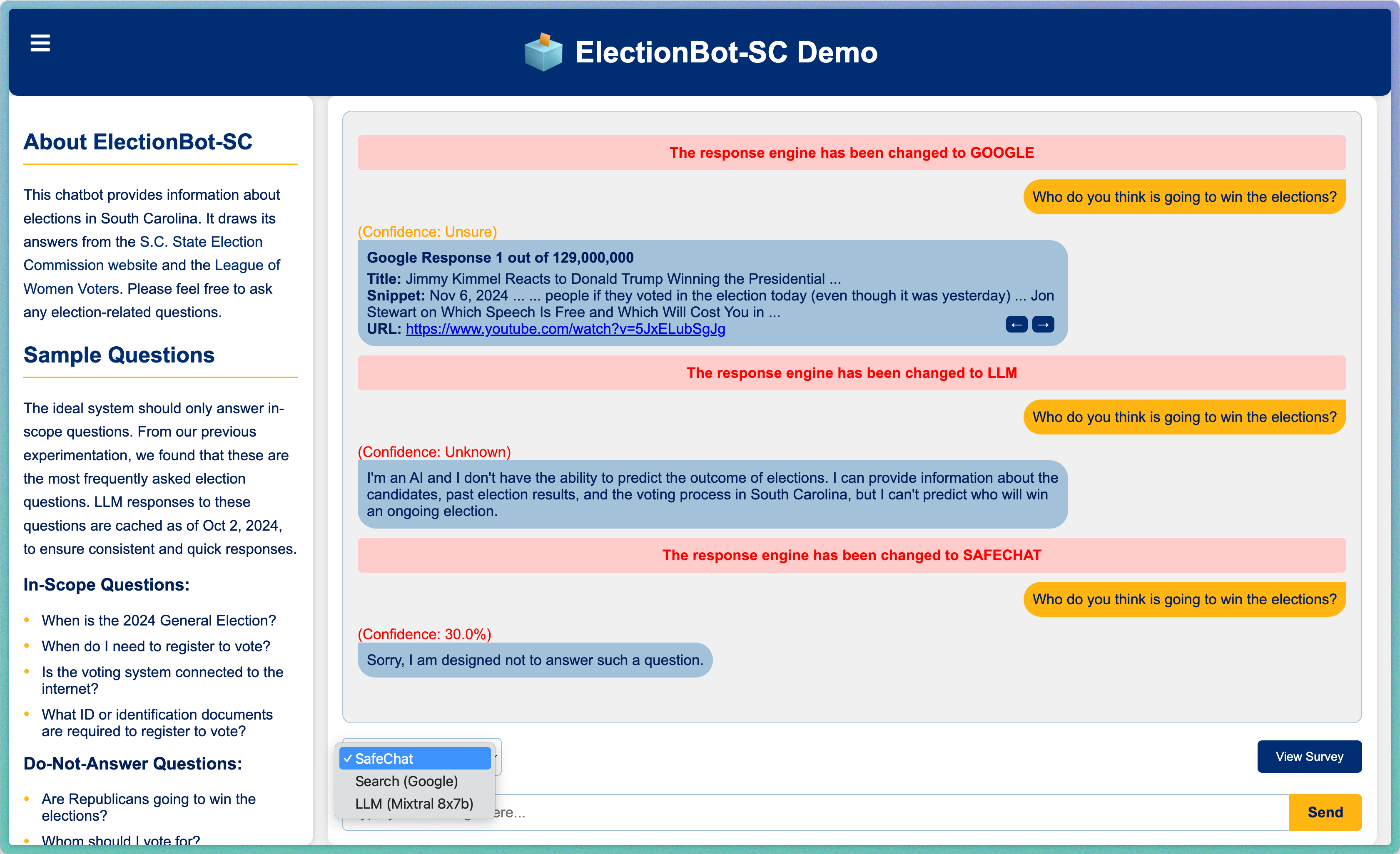}
	\caption{A screenshot of \ebsc demo from 26 March 2025 illustrating user interactions and the ability to switch between response engines. The interface displays color-coded confidence indicators: \sca responses show a numerical score, Google Search is marked ``Unsure'' (orange) but shows result rank, and LLM outputs are labeled ``Unknown'' (red)}.
	\label{fig:eb-sc-demo}
\end{figure*}

\ebsc is implemented as a web application and apart from answering user's SC election related questions, serves as a tool to measure the relative effectiveness of alternative response {\em engines} to provide answers.
The user interface for \ebsc (Figure \ref{fig:eb-sc-demo}) is designed to prioritize usability, transparency, and adaptability, ensuring seamless access to election-related information. The platform integrates \sca as its core response engine and supports interaction with two additional engines, Google Search for real-time web data retrieval and Mixtral 8x7b \cite{jiang2024mixtralexperts}—a free-tier Large Language Model (LLM) offering insights based on pre-trained knowledge.

The interface features a clean and intuitive design with three key components: a collapsible sidebar, a chat area, and a header. The sidebar provides essential information about the chatbot’s purpose, sample queries, and support contacts, helping users understand the scope of the system. The main chat area supports a streamlined interaction flow, displaying conversations with visual cues that indicate both the provenance of responses and their confidence levels. Specifically, \sca responses include a numerical confidence score, color-coded to reflect certainty (green for 0.9 or higher, orange for 0.7--0.9, and red for below 0.7).  Search results are always labeled with an orange ``Unsure'' indicator, reflecting the engine's reliance on real-time data without quality assurance but with rank information of the result shown, while LLM outputs display a red ``Unknown'' label, highlighting the absence of an explicit confidence score. The visual cues  build trust and empowers users to critically assess each engine's output at a glance.

\sca’s rule-based architecture ensures that users receive responses grounded in verified data sources. The interface highlights the provenance tracker in real time, indicating whether responses are sourced from primary, secondary, or other (mixed) sources. This transparency feature, combined with the color-coded confidence indicators, encourages users to engage with \sca’s responses confidently while fostering critical evaluation when alternative engines are used.

The interface also incorporates a survey tool for gathering user feedback on the system’s usability and effectiveness. A dedicated survey button directs users to a concise questionnaire that evaluates satisfaction, relevance of responses, and overall experience. This feedback loop not only aids in continuous improvement but also ensures that the system evolves to meet the dynamic needs of users, particularly during high-stakes periods like elections.

\subsection{Evaluation}

To evaluate the relevance and accuracy of \ebsc in addressing voter information needs, we conducted an Institutional Review Board (IRB)-approved user study \footnote{Research study was certified as exempt
from the IRB per 45 CFR 46.104(d)(3) and 45 CFR 46.111(a)(7) by University of South Carolina IRB\# Pro00124482.} 
The study ran from September 19 - October 4, 2024, a 15 day period  leading to the US  elections (Nov 5, 2024), and had respondents from a University in SC. Due to the    prevailing political environment and to avoid  misunderstanding, we exercised caution by first assuring respondents about the apolitical goals of the study,  and then asking them to participate in the survey.

In the survey, respondents could interact with both  pre-determined election-related  topics whose answers were available in the dataset (and hence by \ebsc), as well as ask on an open-ended topic
\footnote{A pdf version of the survey questionnaire is available at \url{https://shorter.gg/gLkBTk}}. 
The survey included five Election-related question topics (\textbf{EQ}s). Upon shown the system's output, the respondent was asked two User Questions (\textbf{UQ}s), instructing them to rate the relevance and accuracy of each pair on a scale of 1 (strongly disagree) to 5 (strongly agree). Besides \textbf{UQ}s, there was also an option to provide open ended feedback about the survey.  The specific user questions were:

\begin{itemize}
    \item \textbf{UQ1-1, UQ2-1, UQ3-1, UQ4-1}: Respondents indicated to what extent they agreed that each election question matched their own information needs.
    \item \textbf{UQ1-2, UQ2-2, UQ3-2, UQ4-2}: Respondents rated the perceived accuracy of the answer provided by \ebsc for each election question.
    \item \textbf{UQ5}: Respondents were allowed to ask any election-related question to the chatbot. Then, they rated the accuracy of the answer provided specifically for this question.
\end{itemize}

\begin{figure}
    \centering
    \includegraphics[width=0.8\linewidth]{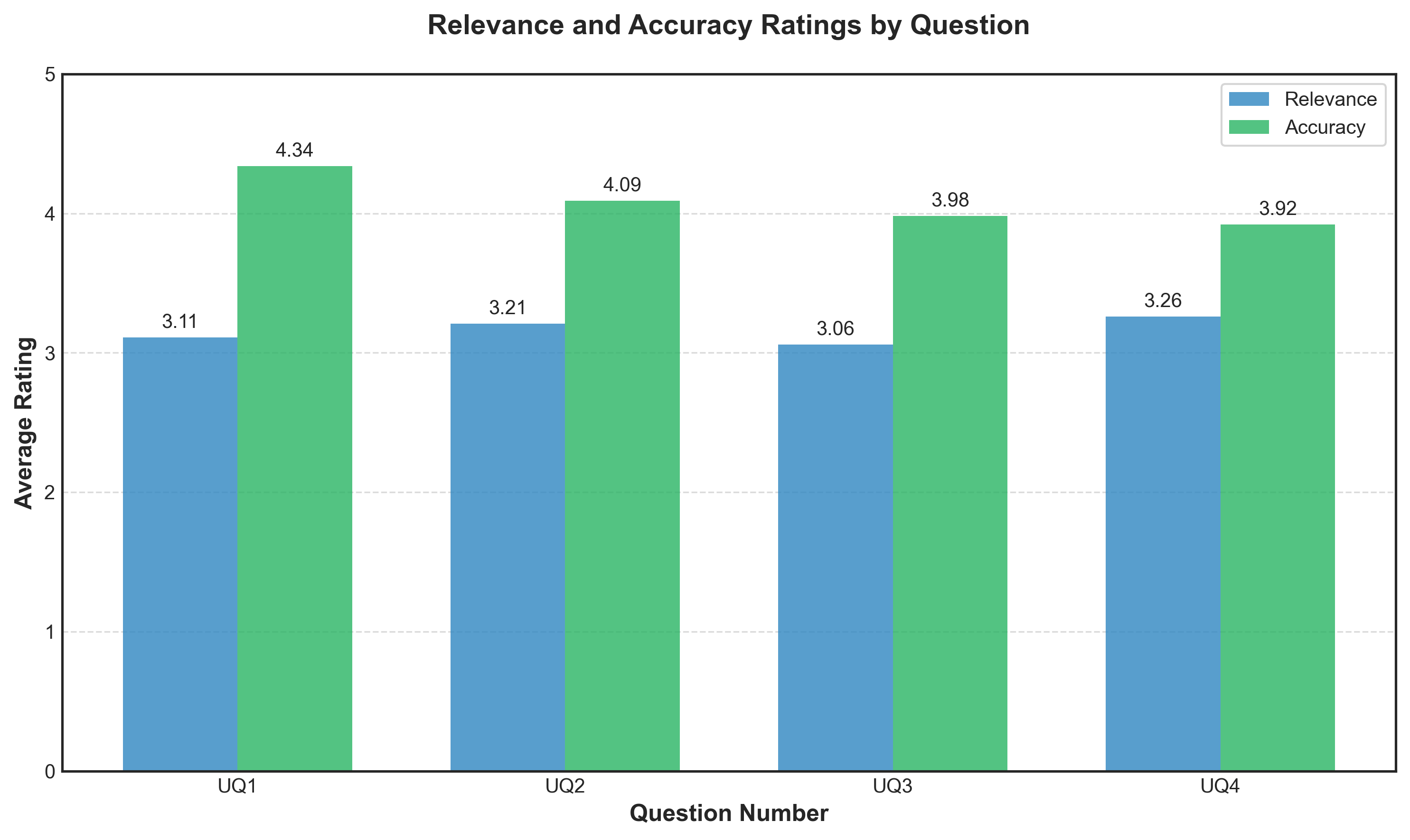}
    \caption{Comparison of relevance and accuracy ratings across different questions in the \ebsc user study. Relevance ratings indicate how well each question matched users' information needs, while accuracy ratings reflect users' perception of answer correctness.}
    \label{fig:relevance_accuracy}
\end{figure}


We received 47 responses to the survey during the period. 
The results for four pre-determined \textbf{EQ}s are  shown in Figure~\ref{fig:relevance_accuracy}. 
We see that the participants generally found the election questions relevant to their information needs, with relevance scores ranging from 3.06 to 3.26 and accuracy from 3.92 to 4.34. All accuracy scores are statistically significant using Chi-square test (p-values $\prec$ 0.05).
The fifth, open-ended election question \textbf{EQ}, which was relevant since the respondent themselves posed it,  
had a lower average accuracy rating of 3.19 (not shown in the figure), indicating that the chatbot's performance may vary depending on the specificity and nature of user questions.





The respondents also gave valuable feedback.  Many respondents appreciated the chatbot's usefulness but suggested enhancing accuracy (e.g., by incorporating more data sources) and source attribution for better reliability. Other enhancements were to provide
more personalized, detailed responses and to improve handling of generic questions. 

The user study, albeit small, is promising. It demonstrates that \ebsc can effectively address voter information needs for  official information. However, performance in handling diverse, user-specific queries requires further optimization. These findings will guide subsequent development and refinement efforts, particularly as we prepare to evaluate the system in new domains and through more rigorous experimental setups, such as randomized controlled trials.

\section{Discussion and Conclusion}
\label{discussion}

The development of \sca, its usage in a number of instances across domains,  and its case study with evaluation in \ebsc, reveals its potential as a scalable and adaptable framework for delivering targeted informational support to users.
We now discuss  \sca from the perspective of an emerging risk guideline. Then, we explore how it may be used within elections to other regions, in different domains, and how the architecture itself can be extended.


\subsection{Assessing \sca on Risk Considerations}
The National Institute of Standards and Technology (NIST) introduced a Risk Management Framework (RMF) that provides a flexible 7-step process for managing information security and privacy risks, aligned with NIST standards and Federal Information Security Modernization Act (FISMA) requirements \cite{nist-rmf}. In \cite{clearview-nist}, the authors  proposes how the NIST RMF can be applied to AI facial recognition software and,  outlines some procedures for ensuring ethical compliance. They introduce an AI System Risk Categorization Matrix (ASRCM) that enables organizations to quickly identify relevant aspects of AI systems, and thereby map and prioritize risks, particularly in contexts where the AI system's impact is extensive. They also propose nine steps that, when followed, ensure the responsible and ethical use of any technology.  As their approach is AI neutral, we apply it to \sca. Table \ref{tab:asrcm} shows the result of applying ASRCM; the application of the nine  steps to \sca is as follows:
\begin{enumerate}

    \item \textbf{Evaluate the ethicality and legality} of the client's purpose in using Chatbot. This determines whether the use justifies the technology's deployment.
    
    \item \textbf{Respect privacy} norms and consent mechanisms, especially in data enrollment and sourcing, ensuring there is no unwarranted intrusion into personal spaces. \sca does not collect any personal data.
    
    \item \textbf{Define the scope} of input questions, whether they are targeted or general, and the scope of answers required to respond to them.  In \sca, the chatbot designer chooses domain-specific FAQ scope, and thus is in control.
    
    \item \textbf{Evaluate chatbot response quality} and accuracy, and check potential errors and biases. 
    
    \item \textbf{Assess the potential for abuse/misuse} by the client. In our case, we assume the chatbot designer uses validated FAQ data.
    
    \item \textbf{Scrutinize} the data sources, whether governmental or private, and ensure they are used for legitimate purposes without infringing on protected activities. In our case, we use open data meant for general use by the public.
    \item \textbf{Assess} if other information retrieval means are overly burdensome or if the chatbot is the most practical and convenient option.
    \item \textbf{Establish} clear policies for how data collected through the chatbot is shared, with whom, and for what purposes, maintaining transparency in its usage.
    \item \textbf{Ensure compliance} with all relevant laws and regulations and adhere to ethical standards and best practices in the deployment of chatbot technology.

\end{enumerate}

\begin{table}[ht]
    \centering
    \resizebox{\columnwidth}{!}{%
    \begin{tabular}{l l l}
    \toprule
    \textbf{Attribute} & \textbf{Categories} & \textbf{\sca} \\
    \midrule
    Use case & Description & Providing safe and reliable output to the public \\
    \midrule
    Potential impact & Health/ Social/ Economic & All could apply, as the system is flexible, but the chatbot designer is in control. \\
    \midrule
    Data sources & 
       \begin{tabular}[t]{@{}l@{}}Sensitive/Non-sensitive\\ Restricted/Public\\ Compliant/Non-complaint\end{tabular} 
       & 
       \begin{tabular}[t]{@{}l@{}}Non-sensitive (domain-specific)\\ Public\\ Compliant\end{tabular} \\
    \midrule
    Level of complexity & White/Gray/Black-box & Chatbot responses (white-box), NLU (black-box) \\
    \midrule
    Regulatory requirements & Yes/No & No (does not apply) \\
    \midrule
    Level of autonomy & Low/ Mid/ High & Mid – controlled autonomy via REST interface to fetch dynamic data \\
    \bottomrule
    \end{tabular}
    }
    \caption{AI System Risk Categorization Matrix (ASRCM) 
    applied to \sca.}
    \label{tab:asrcm}
    \vspace{-0.3in}
\end{table}

We note from Table \ref{tab:asrcm}  that in the case of \sca, the chatbot designer of any instance is in control of the data used in  the system's response and thus, the system's impact. Although the chatbot  has a learning-based, blackbox, NLU for detecting user intent, its response is whitebox since it can be traced to data sources. The chatbot has medium autonomy and no known regulatory consideration. 

\subsection{Impact in Election and Other Domains}
Since election processes in the US vary by states, \sca has already been used to create chatbot instances for a few. 
In \cite{muppasani2023safe}, an earlier version of \sca instance was trained using official election FAQs from the South Carolina (SC) and Mississippi (MS). We also sought feedback from two focus groups of senior citizens. They identified five categories of information that users typically seek - about election procedures,  place to vote, time related, candidates, and propositions (issues), but found that both states only provided information from one category (election procedures). The study highlighted the need for technology-based voter assistance, as some senior citizens faced challenges due to vision, hearing, and cognitive impairments. 
Building on this, in \cite{ai4s2024-voterfaqsUS}, we have collated voter FAQs for all 50 states from state election commissions (primary source) and the non-profit (LWV) as secondary source. Additionally, we have trained an instance of \sca for Massachusetts (MA). Details of these instances are available in \cite{safechat-arch-github}.  


In \cite{finance-llm-icaif}, we evaluated widely used LLM-based chatbots, ChatGPT and Bard, and compared their performance with SafeFinance, a \sca instance  trained on personal finance-related FAQs sourced from Discover, Mastercard, and Visa websites. 
Our findings revealed that SafeFinance maintained consistency in its responses without considering the user's name, while both Bard and ChatGPT exhibited discrepancies, providing additional information based on the race and gender implied by the names used. 
We are currently using \sca to build a chatbot instance aimed at helping medical students train to conduct patient interviews, specifically for sensitive conditions such as HIV/AIDS.  As \sca is open-source, there are also external teams using \sca for their domains. We are aware of one using it to provide library services at a large public University.

\subsection{Extending \sca}

\sca can be extended along many directions. One direction is to increase the automation of  chatbot evaluation  process from surveys to  support for  randomized controlled trial (RCT) - e.g., creation of control and treatment chatbot instances, and analysis of  results. Another direction is to bridge the limitations of LLM-based and rule-based chatbots by creating hybrid  instances guided by  risk consideration that combine the two approaches. A third direction would be to provide support for multiple languages  and content modalities overcoming current limitations of English and text, respectively.


 \section{Conclusion}
The \sca architecture demonstrates the feasibility of quickly creating chatbots, reusing ris-aware best practices, to address trust-sensitive informational needs. We demonstrated that its design enables applications across various domain, offering a promising foundation for advancing user-centered informational support. 

 \section{Acknowledgments}
This work was supported, in part, by faculty awards from Cisco Research and JP Morgan Research, and seed funding from University of South Carolina. Results presented used the Chameleon testbed supported by the National Science Foundation. 


\bibliographystyle{ACM-Reference-Format}
\bibliography{references}



\end{document}